\documentclass{article}

\usepackage{PRIMEarxiv}
\usepackage[colorlinks=true,linkcolor=black,citecolor=blue,urlcolor=blue,]{hyperref}
\usepackage[doipre={doi:~}]{uri}
\RequirePackage[authoryear,round]{natbib}
\usepackage[utf8]{inputenc} 
\usepackage[T1]{fontenc}    
\usepackage{hyperref}       
\usepackage{url}            
\usepackage{booktabs}       
\usepackage{amsfonts}       
\usepackage{nicefrac}       
\usepackage{microtype}      
\usepackage{lipsum}
\usepackage{fancyhdr}       
\usepackage{graphicx}       
\usepackage{xcolor}         
\usepackage{amsmath}
\usepackage{multirow}
\usepackage{booktabs}
\usepackage{amssymb}
\usepackage{bm}
\usepackage{authblk}
\usepackage{subfigure}
\usepackage{hyperref}
\hypersetup{
    colorlinks=true,
    linkcolor=blue,
    filecolor=magenta,      
    urlcolor=cyan,
    pdftitle={Overleaf Example},
    pdfpagemode=FullScreen,
    }
\graphicspath{{media/}}     

\title{\textsc{Reprint}: A randomized extrapolation based on principal components for data augmentation
}
\author{Jiale Wei$^a$, Qiyuan Chen$^a$, Pai Peng$^a$, Benjamin Guedj$^c$, $^*$Le Li$^{a,b}$\\
$^a$ School of Mathematics and Statistics, Central China Normal University, China.\\
$^b$ Hubei Key Laboratory of Mathematical Sciences, China.\\
$^c$ Inria, France \& UCL, United Kingdom\\
Email: leli@mail.ccnu.edu.cn}

\date{} 

\begin{document}

\maketitle
\begin{abstract}

Data scarcity and data imbalance have attracted a lot of attention in many fields. Data augmentation, explored as an effective approach to tackle them, can improve the robustness and efficiency of classification models by generating new samples. This paper presents \textsc{Reprint}, a simple and effective hidden-space data augmentation method for imbalanced data classification. Given hidden-space representations of samples in each class, \textsc{Reprint}~extrapolates, in a randomized fashion, augmented examples for target class by using subspaces spanned by principal components to summarize distribution structure of both source and target class. Consequently, the examples generated would diversify the target while maintaining the original geometry of target distribution. Besides, this method involves a label refinement component which allows to synthesize new soft labels for augmented examples. Compared with different NLP data augmentation approaches under a range of data imbalanced scenarios on four text classification benchmark, \textsc{Reprint}~shows prominent improvements. Moreover, through comprehensive ablation studies, we show that label refinement is better than label-preserving for augmented examples, and that our method suggests stable and consistent improvements in terms of suitable choices of principal components. Moreover, \textsc{Reprint}~is appealing for its easy-to-use since it contains only one hyperparameter determining the dimension of subspace and requires low computational resource.      
\end{abstract}


\section{Introduction}
Deep learning models have achieved state-of-the-art performance on a wide range of supervised learning tasks such as text classification \citep{MKC2021,JHC2020, YDY2019, WSM2018}. However, since these models rely heavily on the use of abundant well-labeled datasets, their applicability may be constrained in some applied settings where labeled data is not only limited but also imbalanced (\citealp{GSM2018}). For example,  with the emergence of new intents or topics in the real world, the virtual conversational assistant is asked to be able to classify these new classes, but the collection of related data for training lags behind \citep{,DAR2021,BTH2017} and their volume is quite small in comparison with that of the old mature classes. 

\begin{figure}[htb]
\centering
\subfigure{\includegraphics[width=8cm,height=5.2cm]{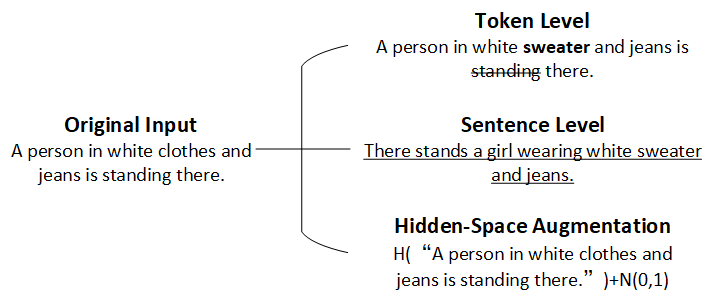}}
\hspace{1mm}
\subfigure{\includegraphics[width=8cm,height=4.8cm]{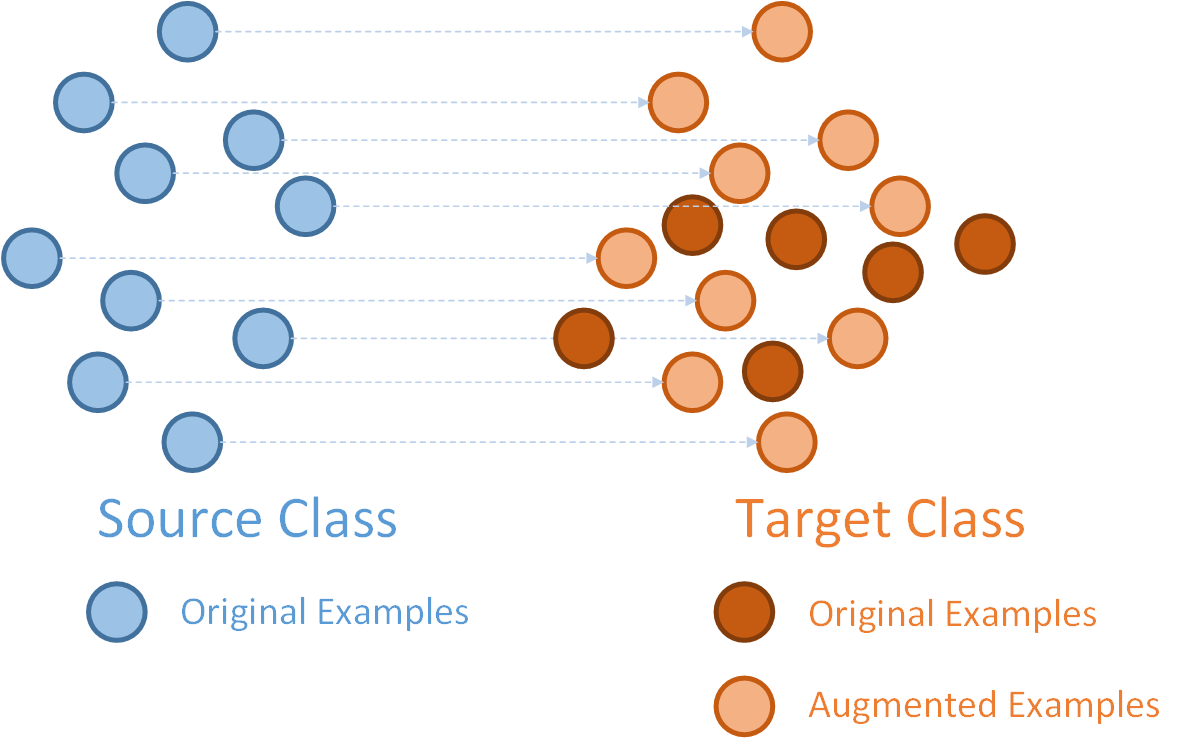}}
\caption{The left figure gives simple examples showing three ways of data augmentation in NLP where $\mathrm{H}(\cdot)$ signifies the hidden-space representation of text and $\mathrm{N(0, 1)}$ the standard Gaussian distribution. The right figure presents how data augmentation method GE3 generates augmented examples.} 
\label{fig:fig_preliminary}
\end{figure}

Data augmentation is a popular method to alleviate data scarcity or imbalance. It refers to generating additional samples to the existing datasets by either duplicating examples of datasets or using transformations or model-based methods to create new synthetic examples \citep{BKR2021, FGW2021, SK2019}. Data augmentation is first widely explored in the computer vision community, where label invariant transformations such as rotation, flipping and cropping have been applied to the original images to increase their amount in classification \citep{TN2018,JWJ2017,KSH2012}. Other methods include noise injection (\citealp{MFJ2018}), mixup (\citealp{ZCD2018}), feature space transfer (\citealp{LWD2018}), random erasing (\citealp{ZZK2020}) and adversary training (\citealp{GSS2015}). The discrete nature of text data along with its complex syntactic and semantic structures make data augmentation a more difficult and underexplored task in Natural Language Processing (NLP). There exists, however, a variety of methods for text data augmentation, and many studies \citep{hu_learning_2019,wei_eda_2019} have shown that these methods are capable of effectively improving model performance especially when the amount of training data is small. In the face of data with imbalanced distribution of categories, text data augmentation aims at expanding samples in minority, thereby reducing the imbalance between categories and improving the generalization ability of models. Its effectiveness has also been proved in practice \citep{ACG2020,wang_thats_2015}. Methods for data augmentation in NLP can be broadly categorized as token-level, sentence-level and hidden-space augmentation (see Figure \ref{fig:fig_preliminary}). Token-level augmentation methods generate new data whose semantic slightly deviates from the original data, based on properly operating on words or phrases of texts. It involves token substitution \citep{wei_eda_2019,Cou2018,XWL2017}, swapping \citep{XH2020, RMH2020,LCD2018}, deletion \citep{RMH2020,wei_eda_2019}, insertion \citep{ MLW2020,wei_eda_2019} and their ensemble (\citealp{And2020}). Instead of modifying tokens, sentence-level augmentation is allowed to change the entire sentence at once while preserving the meaning of the original sentence. It overcomes the limitation of replacement range and increases the diversity of augmented data. One common method for sentence-level augmentation is back-translation which firstly translates original text into other language and then translates it back to obtain augmented data \citep{TSY2021, XDH2020, EOA2018}. Another one is text generation in which new texts can be created via some trained models \citep{ACG2020,Kum2020,MLY2020, HLC2018}. Although back-translation and text generation are able to create novel and diverse data which may not exist in the original dataset, they require either significant training effort or large balanced datasets to fine-tune the model. Being simple yet effective, hidden-space augmentations demand low-resource to produce augmented data since they can work directly on the hidden-space representations of texts by adding noises, performing interpolations \citep{mixtext,2002SMOTE} or extrapolations \citep{Wei2021,VHC2019, DT2017} between text representations.        

\begin{figure}[htb]
\centering
\subfigure{\includegraphics[scale=.3]{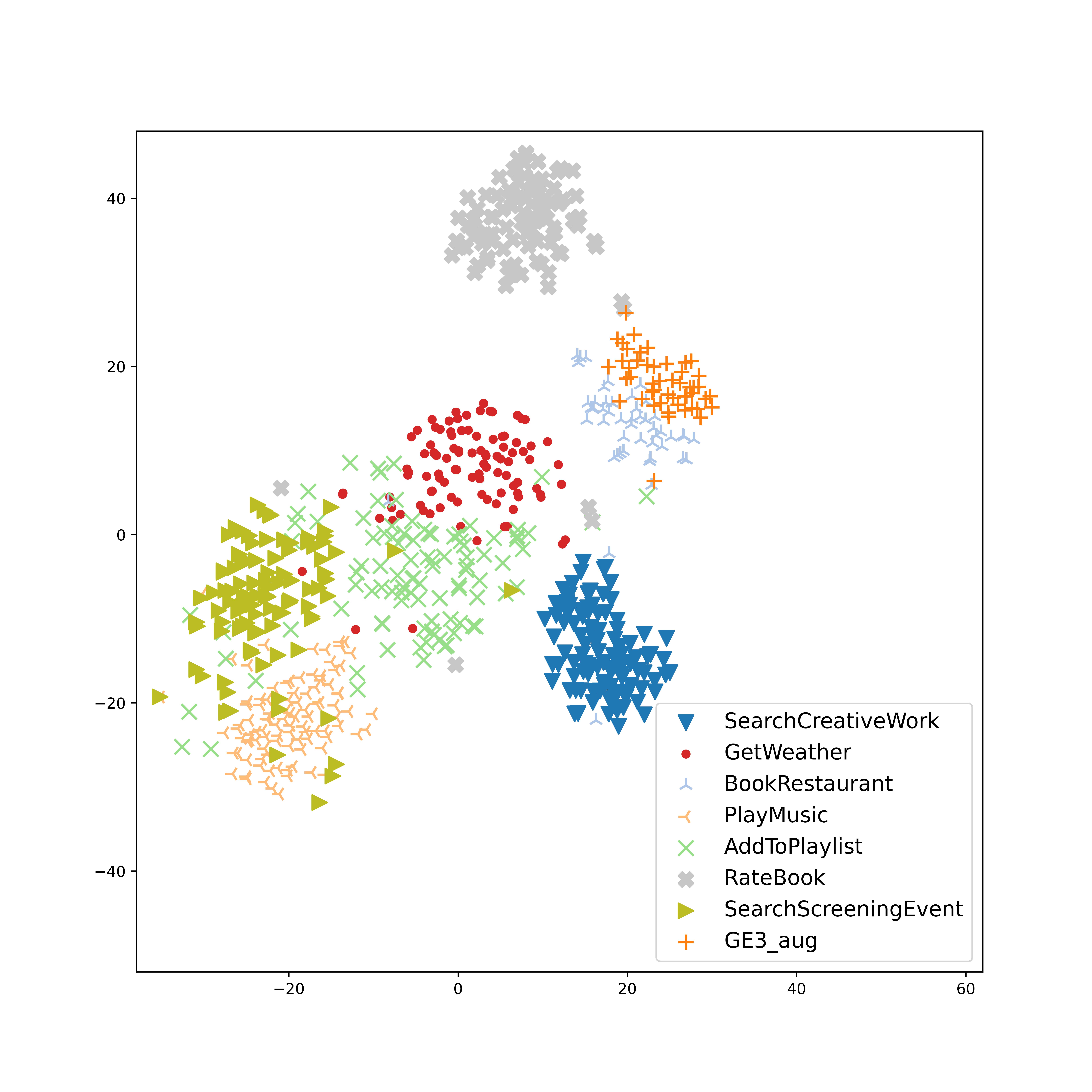}} 
\subfigure{\includegraphics[scale=.3]{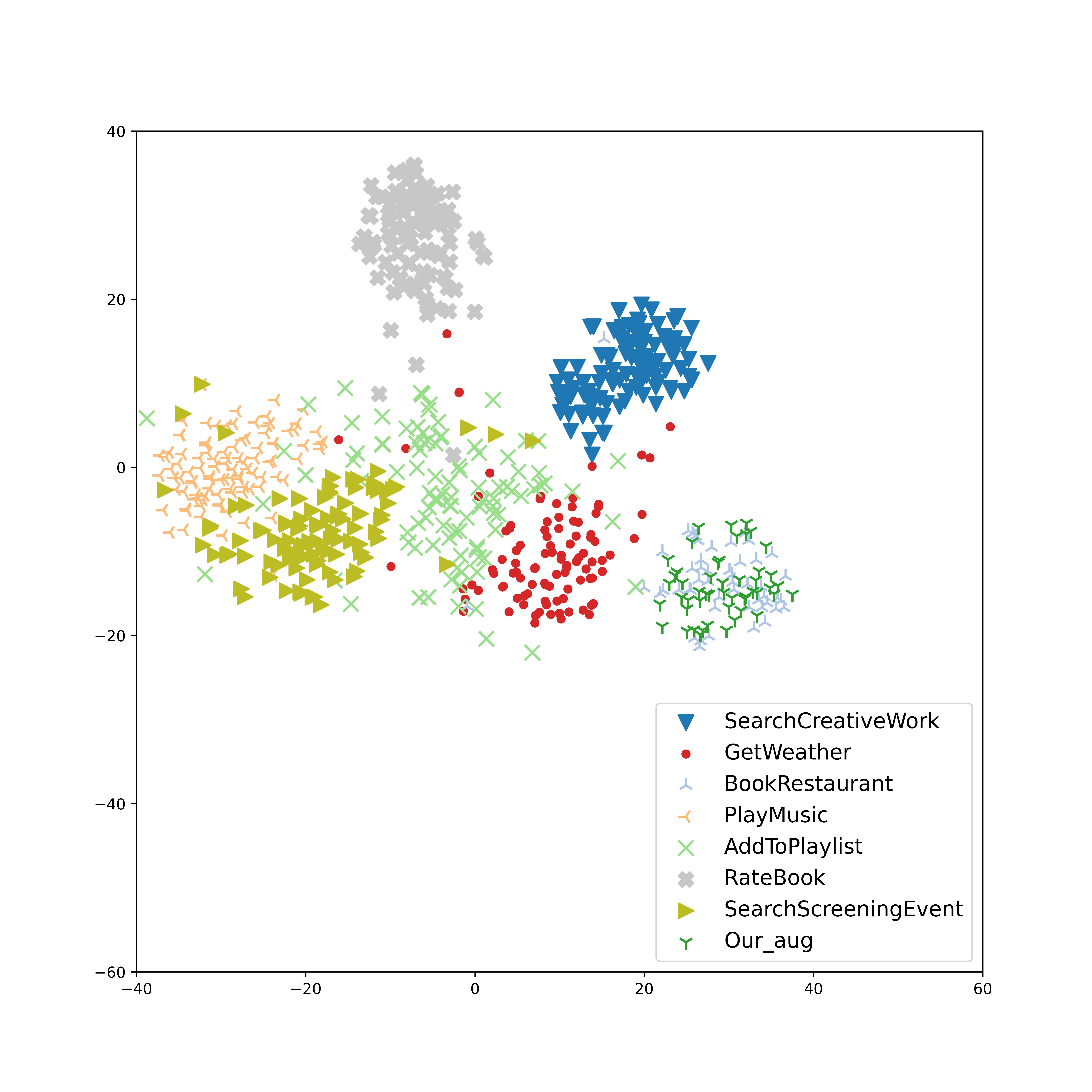}}
\vspace{-2em}
\caption{The t-SNE (\citealp{MH2008}) visualisation of both original and augmented examples produced respectively by GE3 (left panel) and \textsc{Reprint}~(right panel). The source class (RateBook) is in grey, and the target class (BookRestaurant) in light blue. The augmented samples (orange cross) produced by GE3 have a different distribution with respect to the target while that (dark green triangle\_down) created by our method do not.} 
\label{fig:fig1}
\end{figure}

Recently, \cite{Wei2021} proposed a simple data augmentation method called \emph{good-enough example extrapolation} (GE3) and has illustrated its performance in class-imbalanced scenario. It extrapolates the distribution of points in the same category, which models some random variables, from one category onto another (see Figure \ref{fig:fig_preliminary}). More precisely, given hidden-space representations of examples in both majority source and minority target class, GE3 generates augmented examples for the target from the source class by moving examples in the source to the target such that the difference between augmented ones to the target mean equals to that between source examples and their mean. It is shown that it exceeds upsampling and other hidden-space data augmentation methods on three text classification datasets. Despite being model-agnostic, hyperparameter-free and easy-to-use, GE3 fails to consider the geometrical distinction (\citealp{VHC2019}) between distributions of representations in different classes since it simply "copies" the source distribution to the target, which might depart target's original distribution (\emph{e.g.,} Figures \ref{fig:fig_preliminary} and \ref{fig:fig1}). In addition, recent works have revealed that contextualized representations yielded by pretrained language models, BERT (\citealp{bert2019}) for example, contain geometry which is intrinsically low-dimensional and anisotropic \citep{ATV2021, HA2021, PZA2021, Eth2019, GHT2019}. 

In this paper, we aim to take into account spatial patterns between source and target distribution when generating augmented data for imbalanced datasets. We present a simple and effective method for data augmentation, which we call randomized extrapolation based on principal components or \textsc{Reprint}. To incorporate more spatial information of feature embedding distribution between classes, \textsc{Reprint}~firstly uses two subspace generated by principal components \citep{Hot1933} to summarize respectively patterns and trends of both source and target class, then the residual of the projection of each source example to the source subspace is calculated, and finally generates augmented examples for target class by adding residuals from the source to the projection of some random candidate target example to the target subspace. Consequently, the augmented examples will locate around the subspace representation of the target, in other words, they are in consistency with the pattern of the target distribution. 

In addition, we give a strategy to select candidate target examples in \textsc{Reprint}~such that generated examples would be uniformly distributed around the target distribution. We also integrate \textsc{Reprint}~with another component called label refinement, which allows to further create new soft labels for augmented examples. It is motivated by the label synthesis in data augmentation methods such as MIXUP, but has been extended to our extrapolation method. Ablation experiments have demonstrated that it can also boost the performance, serving as an important component for our method.   

We test \textsc{Reprint}~on several commonly used text classification benchmarks in a wide range of class-imbalanced scenarios. Across all benchmarks and scenarios, our method outperforms other eight baseline models. Additionally, by investigating our method with various hyperparameters and hidden-space representations, we demonstrate the stability and universality of it.

The rest of the paper is organized as follows. Section~\hyperref[sec:related work]{2} discusses related work. Section~\hyperref[sec:main method]{3} presents the details of our method \textsc{Reprint}. We illustrate its performance on several text datasets and show the ablation studies in Section~\hyperref[sec:experiments]{4}. Finally, conclusion and future work are given in Section~\hyperref[sec:future work]{5}.


\section{Related work}\label{sec:related work}
\textbf{Hidden-space representation.} Being widely used in NLP community, hidden-space representation are low-dimensional continuous representations of words or sentences that capture their semantic and syntactic. 
\cite{1301.3781} first proposed CBOW and Skip-gram models to construct associations between contexts and central words, which allows central words to obtain their vector representations through contexts. On this basis, \cite{pennington2014glove} further solved the problem of fixed sliding window size and achieved better performance on different tasks.
However, the static word representation produced by the above models cannot solve the problem that word meaning varies according to their context. To address this shortcoming, \cite{2018Deep} and \cite{radford2018improving} pre-trained respectively Bi-LSTM and Transformer-based models on large-scale corpus to produce contextualized word representation which effectively alleviates the problem of polysemy. Both models use an unidirectional network structure, so the relations in the language considered by the current model are incomplete. Furthermore, \cite{bert2019} proposed a Bi-directional Transformer encoder (BERT) to solve this problem. Thanks to this bidirectional structure and two new pre-training tasks (\emph{i.e.,} Masked Language Model and Next Sentence Predication), BERT can learn richer semantic and syntactic information in different layers. For example, \cite{jawahar-etal-2019-bert} found that its intermediate layers encode a rich hierarchy of linguistic information, with surface features at the bottom, syntactic features in the middle and semantic features at the top. And this enables BERT to achieve state of the art performance not only on text classification such as sentiment analysis, natural language inference but also on several other natural language processing tasks.

\textbf{Text data augmentation.} There is a large body of research on text data augmentation \citep{mixtext,wei_eda_2019}. For example, word or phrase replacements which transform parts of the original texts are used for text augmentation in classification \citep{XDH2020,wei_eda_2019}. More advanced approaches which are capable of generating whole new texts depend on back-translation or retraining of specific neural models. These methods, however, either require a large number of training data to generate good example that would improve down-stream tasks or demand for very high computational cost. Hidden-space augmentation, which is the most related to our work, seeks to overcome these challenges by performing deformations of data examples in hidden space rather than at text-level. \cite{ZCD2018} introduced a simple and model-agnostic data augmentation routine, termed MIXUP, to construct augmented training examples based on the convex combinations of pairs of examples and their labels. \cite{mixtext} later extended this interpolation-based routine to semi-supervised text classification and selected beta distribution as the sampling distribution for the ratio to trade off between pairs of examples. In parallel, extrapolation-based and label preserving methods have also been proposed for data augmentation. \cite{DT2017} proposed to synthesize new examples for a given class by extrapolating a pair of adjacent points within the same class. Following a similar idea, \cite{Kum2020} studies several hidden-space augmentation techniques to improve few-shot intent classification. \cite{Wei2021} presented GE3 which further extends extrapolation from between points in the same class to distributions of different classes, and showed its performance in imbalanced class setting. Although inspired by the former, our method is notably different: firstly, the underlying geometry of distribution between classes is ignored in prior work whereas \textsc{Reprint}~uses subspace spanned by principal components to characterize it. Secondly, \textsc{Reprint}~is also able to synthesize new labels, rather than just hidden-space representations, for augmented examples.

\textbf{Label smoothing.} It is first proposed by \cite{2016Rethinking} in order to provide regularization for deep neural networks. Instead of using hard (\emph{i.e.,} one-hot) labels for training, \cite{2016Rethinking} introduced label smoothing,which improves accuracy by computing cross entropy not with the "hard" targets from the dataset,but with a weighted mixture of these targets with the uniform distribution. Although such soft labels can be effective in giving strong regularization to model while avoiding overfitting, they treats each category equally and does not take into account the importance of target category.
In the field of NLP, label smoothing approach has also proved its effectiveness in practice \citep{2021Similarity,2020Semantic}.
Further, \cite{1906.02629} meticulously analyzed the principles of label smoothing and explained several behaviors (\emph{e.g.,} closer sample cluster) observed when training deep neural networks with label smoothing. They also found that label smoothing impairs distillation, despite the positive effect on generalization and calibration. To obtain a more reasonable probability distribution between the target and non-target categories, \cite{2021Delving} proposed a new label smoothing strategy, called Online Label Smoothing, this strategy generates soft labels based on model prediction statistics of the target category. Experiments have also demonstrated the effectiveness of this approach on classification tasks. Alternatively, the mixup strategy can be seen as a method of label smoothing \citep{mixtext,ZCD2018}. This method achieves the effect of label smoothing by weighting the labels of two training samples by averaging them, which makes the decision boundary obtained from training smoother and effectively improves the generalization performance of the model.

\section{Introducing \textsc{Reprint}}\label{sec:main method}
In this section, we introduce our method, including its formal expression, the choice of principal components to form subspace representation to summarize class distribution, the randomized strategy to select candidate examples needed in the method, and the soft labels to be endowed with the augmented examples. 

\subsection{Randomized extrapolation based on principal components}
Given a text classification task with $K$ output classes $\{c_{k}, k=1,2,\dots, K\}$, let us denote by $\{X^{c}_{i} \in \mathbb{R}^{d}, i=1, 2, \dots, n_{c}\}$ the hidden-space representations of $n_{c}$ training examples in class $c$. For each class $c$, the mean $\mu_{c}$ of all hidden-space representations in that class, and the centralized version $\widetilde{X}^{c}_{i}$ of each representation $X^{c}_{i}, i=1, 2, \dots, n_{c}$ are defined respectively as    
\begin{equation*}
   \mu_{c}=\frac{1}{n_{c}}\sum_{i=1}^{n_{c}} X^{c}_{i}, \quad \quad  \widetilde{X}^{c}_{i}=X^{c}_{i} -\mu_{c}. 
\end{equation*}
In addition, let $\{\alpha_{i}^{c}, i=1, 2,\dots, h\}$ denote the first $h$ principal components of class $c$, built on $\{\widetilde{X}_{i}^{c}, i=1,2,\dots, n_{c}\}$, and $S_{h}^{c}$ the subspace spanned by these $p$ principal components. The projection of centered representation $\widetilde{X}^{c}_{i}$ to $S_{h}^{c}$ is given by
\begin{equation}\label{eq:projection expression}
    \text{Proj}_{S_{h}^{c}}\left(\widetilde{X}^{c}_{i}\right) = A_{h}^{c}\left(\left(A_{h}^{c}\right)^{t}\widetilde{X}^{c}_{i}\right),
\end{equation}
where $A_{h}^{c}=[\alpha_{1}^{c}, \alpha_{2}^{c},\dots, \alpha_{h}^{c}]$ is the projection transformation matrix whose columns are principal components, and $A^{t}$ represents the transpose of matrix $A$. 

\textsc{Reprint}~attempts to generate, in a randomized fashion, augmented examples by extrapolating the data distribution from a source class $c_{s}$ to the target $c_{t}$ in hidden space. More precisely, for each  $\widetilde{X}^{c_{s}}_{i}$ in the source class, a corresponding augmented example $\widehat{X}^{c_{t}}$ in target class is generated as follows
\begin{equation}\label{eq:randomized extrapolation on pc}
    \widehat{X}^{c_{t}} = \widetilde{X}^{c_{s}}_{i} - \text{Proj}_{S_{h}^{c_{s}}}\left(\widetilde{X}^{c_{s}}_{i}\right) + \text{Proj}_{S_{q}^{c_{t}}}\left(\widetilde{X}^{c_{t}}_{J}\right) + \mu_{c_{t}},
\end{equation}
where $\widetilde{X}^{c_{t}}_{J}$ is a random candidate example to be sampled from a distribution $P$ whose support lies on the centered training set $\left\{\widetilde{X}^{c_{t}}_{j}, j=1,2,\dots, n_{c_{t}}\right\}$ of target class, and the third term on the right hand side (R.H.S) of equation \eqref{eq:randomized extrapolation on pc} is the projection of $\widetilde{X}^{c_{t}}_{J}$ to the subspace $S_{q}^{c_{t}}$ formed by the first $q$ principal components of target class. It can be noted that augmented example is build upon three parts: the residual of the projection of a source example, the projection of a random target example and the target mean. Therefore, while increasing the amount and diversity of target class, the augmented examples generated remain to be around the subspace representation of target class, hence making them consistent with the original geometry of target distribution. In addition, we can also see augmented examples from another perspective: subspace representations of both source and target class can be interpreted as conserving primary and common semantic and syntactic structure shared by the examples within each class, and the residual serves as a variant of the main structure contained in the source class. Adding this variant to the projection of a target example may therefore help to diversify the augmented example while preserving similar meaning with the origin. 

For each class in the training set, we could generate a set of extrapolated samples from every other class, multiplying approximately the size of the training set by a factor $K$. In the training process, the classification model will be trained on the union of original data and extrapolated ones. It is worth noticing that \textsc{Reprint}~is model-free and simple to use, requiring little computation cost since the only hyperparameters involved are the number $h, q$ of principal components for both source and target class.

\subsection{Choice of principal components and sampling distribution}\label{sec:choice of pc}
We propose first two heuristic strategies to determine the number $h$ and $q$ (\emph{i.e.,} the dimensionality of subspaces) of principal components used to explain the source and target class. The first one is to consider two subspaces of same dimension (\emph{i.e.,} $h=q$) to represent both source and target distribution, and choose an optimal value in terms of its performance on test set. The second one relies on the explained variance ratio which is defined as the percentage of variance explained by each of the selected components, and $h$ (or $q$) is selected such that a certain amount, for example 90$\%$, of variance can be explained by the first $h$ (or $q$) principal components in source (or target) class.

In addition, for the choice of sampling distribution $P$ for random candidate $\widetilde{X}^{c_{t}}_{J}$ in equation \eqref{eq:randomized extrapolation on pc}, we set it as uniform distribution over centered target training set $\left\{\widetilde{X}_{c_{t}}^{j}, j=1,2,\dots n_{c_{t}} \right\}$ of target class $c_{t}$, \emph{i.e.,}
\begin{equation*}
    P\left(\widetilde{X}_{J}^{c_{t}}=\widetilde{X}^{c_{t}}_{j}\right) = \frac{1}{n_{c_{t}}}, \, j=1, 2,\dots, n_{c_{t}}.
\end{equation*}
Following this choice, augmented examples generated by equation \eqref{eq:randomized extrapolation on pc} will be uniformly distributed along the underlying pattern of target distribution.

\subsection{Label refinement for augmented examples}\label{sec:label refinement}
We also propose another component named label refinement which attempts to integrate label synthesis used in interpolation-based augmentation into our method to generate new labels, rather than just hidden-space representations, for augmented examples. By combining equations \eqref{eq:projection expression} and \eqref{eq:randomized extrapolation on pc}, the augmented example $\widehat{X}^{c_{t}}$ for target class $c_{t}$ can be rewritten as 
\begin{align*}
    \widehat{X}^{c_{t}} &= \left(I_{d} - A_{h}^{c_{s}}\left(A_{h}^{c_{s}}\right)^{t}\right) \widetilde{X}^{c_{s}}_{i} + A_{q}^{c_{t}}\left(A_{q}^{c_{t}}\right)^{t} \widetilde{X}^{c_{t}}_{J} + \mu_{c_{t}} \\
    & \triangleq W^{c_{s}}\widetilde{X}^{c_{s}}_{i} + W^{c_{t}}\widetilde{X}^{c_{t}}_{J} + \mu_{c_{t}},
\end{align*}
where $I_{d}$ represents the identity matrix. We synthesize new label $\hat{y}^{c_{t}}$ for $\widehat{X}^{c_{t}}$ as
\begin{equation}\label{eq:created label}
\hat{y}^{c_{t}}=\left\{
\begin{aligned}
&\lambda y^{c_{s}} +(1-\lambda) y^{c_{t}}, && \text{if}\,\, |W^{c_{s}}| >0 \,\, \text{and} \,\, |W^{c_{t}}|>0, \\
&y^{c_{t}}, && \text{else}.
\end{aligned}
\right.
\end{equation}
where $y^{c_{s}}$ and $y^{c_{t}}$ denotes respectively one-hot label (\emph{i.e.,} hard label) representation for class $c_{s}$ and $c_{t}$, and $\lambda=|W^{c_{s}}|/\left(|W^{c_{s}}| + |W^{c_{t}}|\right)$ 
represents the ratio between two determinants $|W^{c_{s}}|$ and $|W^{c_{t}}|$. In other words, if the augmented example is extrapolated as the sum of two linear transformations for which inner products between $W^{c_{s}}\widetilde{X}^{c_{s}}_{i}$ (\emph{respectively}  $W^{c_{t}}\widetilde{X}^{c_{t}}_{J}$) and its domain $\widetilde{X}^{c_{s}}_{i}$ (\emph{respectively} $\widetilde{X}^{c_{t}}_{J}$) are both strictly positive, then the new label for augmented example is created as a linear combination of both source and target label where the weight is determined by the determinant of two transformation matrix. Otherwise, it is preserved as the hard label of the target class.

\section{Experiments}\label{sec:experiments}

In this section, we test \textsc{Reprint}~on various text classification benchmark datasets \footnote{The code is available in: \href{https://github.com/bigdata-ccnu/REPRINT}{https://github.com/bigdata-ccnu/REPRINT}}. We first give a brief introduction about the datasets, the class-imbalanced scenario and experiment settings on which our experiments are performed. Next, we describe some baseline models with which we are about to compare, including some data augmentation baselines relying on extrapolation and other baselines built on either interpolation or token-level augmentation. Finally, we present the performance of both our method and baselines on various datasets, the results of which demonstrate the effectiveness of our approach. Additional ablation studies are also given, showing how different components involved in our method contribute to the result.

\subsection{Datasets}
We run our experiments on the following benchmarks in text classification. 

\begin{enumerate}
    \item \textbf{SNIPS} \citep{CSB2018}: Snips Voice Platform  is a dataset of over 16,000 crowdsourced queries distributed among 7 user intents of various complexity. Each intent has around 1,800 training samples, and the test set has 700 samples altogether.
    \item \textbf{Yahoo Answers} \citep{chang2008importance}: The Yahoo Answers topic classification dataset  consists of 10 largest main categories from the Yahoo! Answers Comprehensive Questions and Answers. Each class contains 6,000 testing samples. As original training set is too large, we constraint its size to at most 3000 for each class.
    \item \textbf{AG News} \citep{2015Character}: It  consists of 4 largest classes of original news corpus. Each topic class has 30,000 training samples and 1,900 testing samples. Due to its large training set, the maximum number of training samples used for each class in training is set to 2,000.
    \item \textbf{DBpedia} \citep{dbpedia}: It includes structured content collected from Wikipedia. For each of the 14 classes, we only use a subset containing 2,500 out of 40,000 training samples, but evaluating our approach on the whole test set.

\end{enumerate}

\subsection{Experiment settings}
For the above datasets, we artificially create class-imbalanced scenario through random sampling. Specifically, half of the classes of each dataset would be randomly selected to equip with only a small number $n_{\textrm{small}}$ of samples while each of the remaining classes would set to have $n_{\textrm{large}}$ samples (\emph{i.e.,} $n_{\textrm{large}}=\{1800, 3000, 2000, 2500\}$ respectively for the four datasets mentioned above). Moreover, we consider a range of values $(\emph{i.e.,} 32, 64, 128)$ for $n_{\textrm{small}}$ to simulate different extends of class-imbalanced scenarios. Besides, throughout the experiments, we use BERT-based-uncased \cite{bert2019} pre-trained language model to encode the text, and represent it in hidden-space using average pooling over the output of encoder's last layer. The weights in BERT encoder are frozen, and an additional MLP classifier with softmax is added on top for classification. After applying data augmentation methods to these imbalanced datasets, the MLP classifier would leverage both original and augmented data for training and the accuracy is obtained on the whole test set. We run all experiments for five random seeds and the encoder only reads the first 512 tokens of text without any additional preprocessing.

\subsection{Baselines}
To test the effectiveness of our method, we compare it with following baseline models:

\begin{enumerate}
    \item \textbf{UPSPL} (Upsampling) is a simple tool handling class-imbalanced scenario by sampling directly from dataset.
    \item \textbf{NOISE} (Gaussian Noise) generates samples from $\mathcal{N}(0,0.1)$ and adds them to the existing samples to form augmented example.
    \item \textbf{SMOTE} \citep{2002SMOTE} is a simple and commonly used method to solve data imbalance. By selecting examples that are close in the feature space using K-Nearest Neighbor, new samples are created by their interpolations.
    \item \textbf{EDA} \citep{wei_eda_2019} is an easy way to perform data augmentation while having high interpretability. It makes good use of text editing techniques and could generate novel samples with a certain degree of diversity.
    \item \textbf{TMIX} \citep{mixtext} creates augmented samples in hidden space. Given a pair of examples in two different classes, augmented features and labels are interpolated by their convex combination. 
    
    \item \textbf{WE} (Within-Extrapolation \citep{DT2017}) extrapolates between hidden-space representations of two examples $X_{i}^{c}$ and $X_{j}^{c}$ of the same class $c$ to create augmented example as
\begin{equation*}
    \widehat{X}^{c} = \lambda (X_{i}^{c}-X_{j}^{c}) + X_{i}^{c},
\end{equation*}
    where $\lambda$ is set to be 0.5 in the experiment.
    \item \textbf{LD} (Linear Delta \citep{VHC2019}) adds the bias between two examples $X_{i}^{c}$ and $X_{j}^{c}$ of the same class $c$ to another sample $X_{k}^{c}$ to form an augmented example as
\begin{equation*}
    \widehat{X}^{c} = (X_{i}^{c}-X_{j}^{c}) + X_{k}^{c}
\end{equation*}
     \item \textbf{GE3} \citep{Wei2021} generates an augmented sample $\widehat{X}^{c_t}$ for target class $c_{t}$ by extrapolating from a sample $X^{c_s}_{i}$ in source class $c_{s}$ as 
    \begin{equation*}
        \widehat{X}^{c_t} = X^{c_s}_{i} - \mu_{c_{s}} + \mu_{c_{t}}
    \end{equation*}
        where $\mu_{c_{s}}$ and $\mu_{c_{t}}$ are respectively the mean of original samples in class $c_{s}$ and $c_t$.

\end{enumerate}

To cope with data imbalanced scenarios, our method would follow the same data augmentation strategy as GE3. As for other approaches mentioned above, they would generate samples for minority classes until that training samples of each class are of approximately same size.

\subsection{Results}

\subsubsection{Test accuracy of models}

Table \ref{tab:table1} compares the performance of \textsc{Reprint}~and baseline models on four text classification datasets with the size $n_{\textrm{small}}$ of minority class varying from 32 to 128. The mean accuracy and standard deviation are calculated based on five random seeds. It is noted that \textsc{Reprint}~consistently outperforms baselines on all benchmarks, with an average improvement of 5.20\% of accuracy over UPSPL under different $n_{\textrm{small}}$ values.

\begin{table}[htb]
\centering
\begin{tabular}{c|ccc|ccc}
\hline
Dataset      & \multicolumn{3}{c|}{Snips}   & \multicolumn{3}{c}{Yahoo Answer}    \\ \hline
$n_{\textrm{small}}$  & 32       & 64       & 128    & 32      & 64      & 128     \\ \hline
UPSPL   & $88.97_{\pm0.83}$     & $92.11_{\pm1.06}$     & $93.60_{\pm0.51}$   & $46.88_{\pm0.39}$   & $50.36_{\pm0.39}$     & $53.84_{\pm0.53}$    \\
SMOTE        & $89.00_{\pm0.83}$     & $91.97_{\pm0.91}$     & $93.54_{\pm0.51}$   & $46.62_{\pm0.43}$    & $50.52_{\pm0.43}$     & $54.12_{\pm0.58}$    \\
TMIX         & $89.00_{\pm1.61}$     & $92.46_{\pm1.07}$     & $94.34_{\pm0.28}$   & $53.92_{\pm3.33}$    & $59.51_{\pm3.22}$     & $60.08_{\pm0.78}$    \\
EDA          & $88.31_{\pm1.22}$     & $92.01_{\pm1.34}$     & $92.65_{\pm0.57}$   & $46.24_{\pm0.34}$    & $50.52_{\pm0.35}$     & $53.78_{\pm0.41}$    \\
LD           & $85.74_{\pm1.09}$     & $89.69_{\pm0.57}$     & $92.11_{\pm0.78}$   & $46.70_{\pm0.45}$    & $52.25_{\pm0.23}$     & $58.88_{\pm0.39}$    \\
WE           & $85.69_{\pm0.97}$     & $90.06_{\pm0.77}$     & $92.43_{\pm0.39}$   & $44.44_{\pm0.53}$    & $49.36_{\pm0.42}$     & $55.87_{\pm0.43}$    \\
NOISE        & $89.17_{\pm0.99}$     & $92.60_{\pm0.93}$     & $94.00_{\pm0.78}$   & $53.47_{\pm0.24}$    & $59.48_{\pm0.41}$     & $63.42_{\pm0.52}$     \\
GE3          & $92.03_{\pm1.19}$     & $94.23_{\pm0.77}$     & $95.03_{\pm0.58}$   & $53.32_{\pm0.24}$    & $59.06_{\pm0.18}$     & $63.04_{\pm0.41}$     \\
\textsc{Reprint}         & $\bm{92.05_{\pm1.15}}$     & $\bm{94.23_{\pm0.90}}$     & $\bm{95.03_{\pm0.36}}$   & $\bm{59.01_{\pm0.47}}$    & $\bm{63.08_{\pm0.26}}$    &    $\bm{65.14_{\pm0.64}}$     \\ \hline
             & \multicolumn{3}{c|}{AG news} & \multicolumn{3}{c}{DBpedia} \\ \hline
UPSPL   & $74.08_{\pm1.73}$     & $79.14_{\pm1.03}$     & $82.56_{\pm0.30}$   & $93.38_{\pm0.18}$    & $95.50_{\pm0.41}$    & $96.85_{\pm0.13}$    \\
SMOTE        & $74.23_{\pm1.74}$     & $79.31_{\pm1.00}$     & $82.63_{\pm0.27}$   & $93.24_{\pm0.20}$    & $95.36_{\pm0.41}$    & $96.73_{\pm0.16}$    \\
TMIX         & $75.64_{\pm1.85}$     & $80.20_{\pm1.23}$     & $83.76_{\pm0.31}$   & $94.12_{\pm0.31}$    & $96.47_{\pm0.55}$    & $97.50_{\pm0.40}$    \\
EDA          & $71.64_{\pm1.87}$     & $77.41_{\pm1.47}$     & $81.62_{\pm0.58}$   & $92.61_{\pm0.35}$    & $94.86_{\pm0.37}$    & $96.45_{\pm0.15}$    \\
LD           & $68.53_{\pm1.71}$     & $73.11_{\pm1.01}$     & $75.83_{\pm0.99}$   & $90.31_{\pm0.14}$    & $93.45_{\pm0.34}$    & $95.33_{\pm0.31}$    \\
WE           & $67.74_{\pm1.67}$     & $72.97_{\pm0.80}$     & $76.44_{\pm0.90}$   & $90.44_{\pm0.21}$    & $93.66_{\pm0.49}$    & $95.75_{\pm0.21}$    \\
NOISE        & $75.64_{\pm1.30}$     & $80.79_{\pm0.81}$     & $84.07_{\pm0.34}$   & $93.67_{\pm0.17}$    & $95.73_{\pm0.40}$    & $97.04_{\pm0.11}$    \\
GE3          & $78.10_{\pm1.46}$     & $82.07_{\pm0.61}$     & $85.19_{\pm0.31}$   & $94.20_{\pm0.30}$    & $95.90_{\pm0.29}$    & $96.94_{\pm0.21}$    \\
\textsc{Reprint}         & $\bm{80.16_{\pm1.04}}$     & $\bm{84.04_{\pm0.86}}$     & $\bm{86.88_{\pm0.32}}$   & $\bm{95.63_{\pm0.16}}$    & $\bm{96.86_{\pm0.28}}$    & $\bm{97.54_{\pm0.12}}$    \\ \hline
\end{tabular}
\vspace{0.7em}
\caption{Mean accuracy (\%) and standard deviation of different models on four text classification benchmarks with various $n_{\textrm{small}}$ values. The mean accuracy and its standard deviation are calculated using five random seeds. The reported results of our approach is obtained with five principal components for every class.}
\label{tab:table1}
\end{table}

In addition, although WE and LD also use extrapolation skill as ours, their comparatively poor performance indicates that their way of in-class extrapolation is not enough to restore the original distribution of target class as our method does. SMOTE outweighs mostly the formal two but still falls behind \textsc{Reprint}, implying that integrating the geometry of class distribution into extrapolation could further increase the quality of augmented examples. At the same time, EDA does not achieve the same accuracy as some of the hidden-space augmentation methods such as GE3 or \textsc{Reprint}. We argue that this may due to that modifications of tokens on the input-level of text cannot result in augmented examples as diverse as produced by other augmentation methods, hence limiting the performance of it. Moreover, if one compares \textsc{Reprint}~with NOISE, the improvement of accuracy shows that the prior information about the distribution structure of classes used in our method do contribute to the performance. 

Moreover, it is worth noticing that both TMIX and \textsc{Reprint}~use soft labels in training. From results in Table \ref{tab:table1}, the performance of TMIX is not as good as ours on three datasets (\emph{i.e.,} Snips, Yahoo Answer and AGnews) and it generally has larger standard deviation in all cases. It may be because the decision boundaries between classes are not strong enough yet, or a single-layer MLP classifier cannot provide stable training like the one \cite{mixtext} did on BERT. At the same time, it suggests that augmented examples generated by our method have smaller perturbation, enabling faster convergence of the classifier.  

Especially, \textsc{Reprint}~achieves comprehensive improvements in respect to datasets and sizes of minority class over one of the best competitor GE3 which can be in fact regarded as a simplified version of our approach. This performance gain demonstrates the advantage of using principal components in our method to not only account for the pattern of distribution of different classes when generating augmented examples, but also to guide the synthesis of new labels for them. The Figure \ref{fig:fig1} leverages t-SNE technique (\citealp{MH2008}) to visualize both original and augmented examples produced respectively by GE3 (left panel) and \textsc{Reprint}~(right panel) in a two-dimensional plane. It illustrates the effectiveness of our method since the augmented examples generated by GE3 deviate from the target class (BookRestaurant) while that generated by ours still have fidelity to the original target distribution.

\begin{figure}[t]
\centering
\subfigure{\includegraphics[scale=.5]{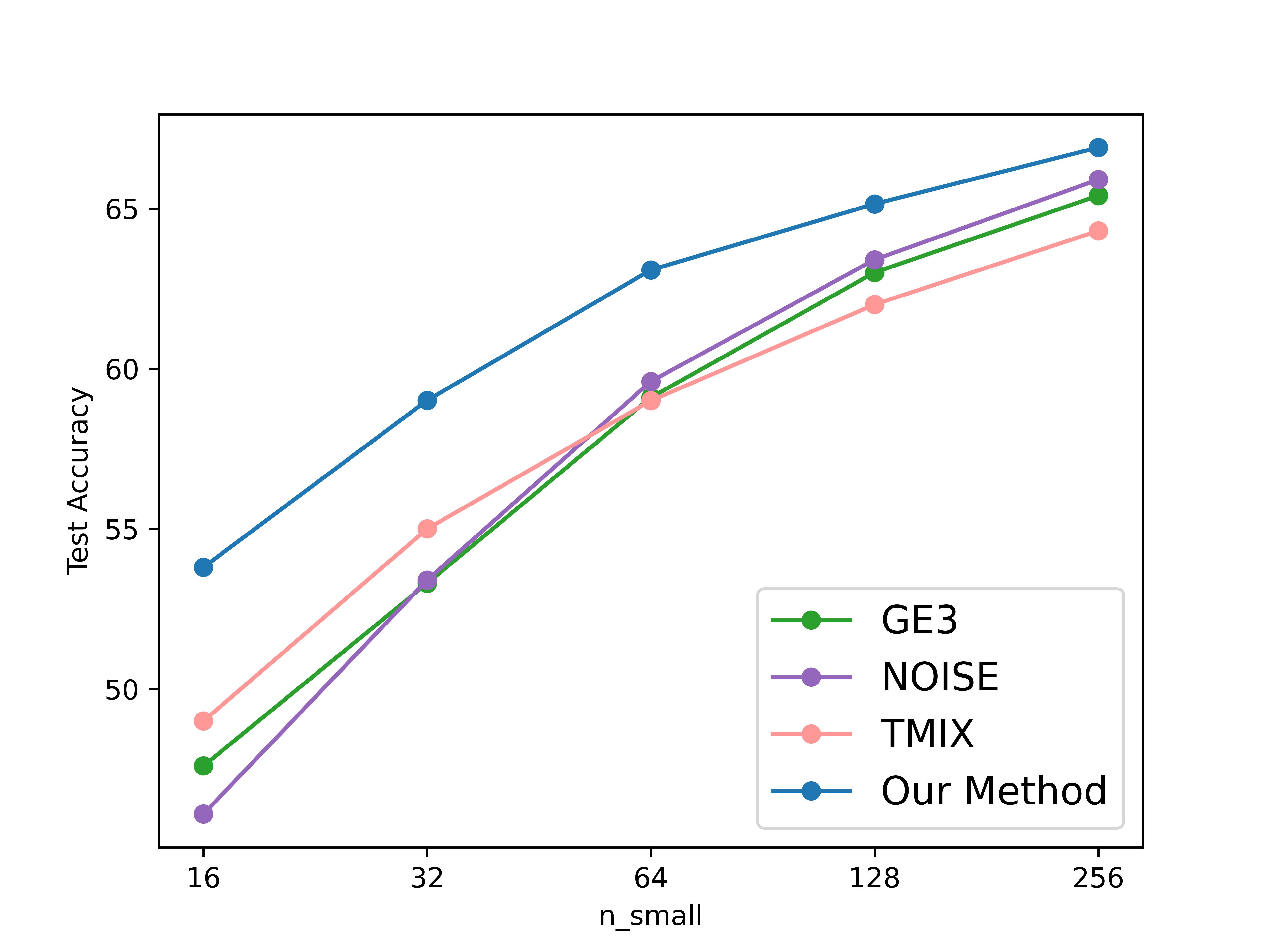}}
\subfigure{\includegraphics[scale=.5]{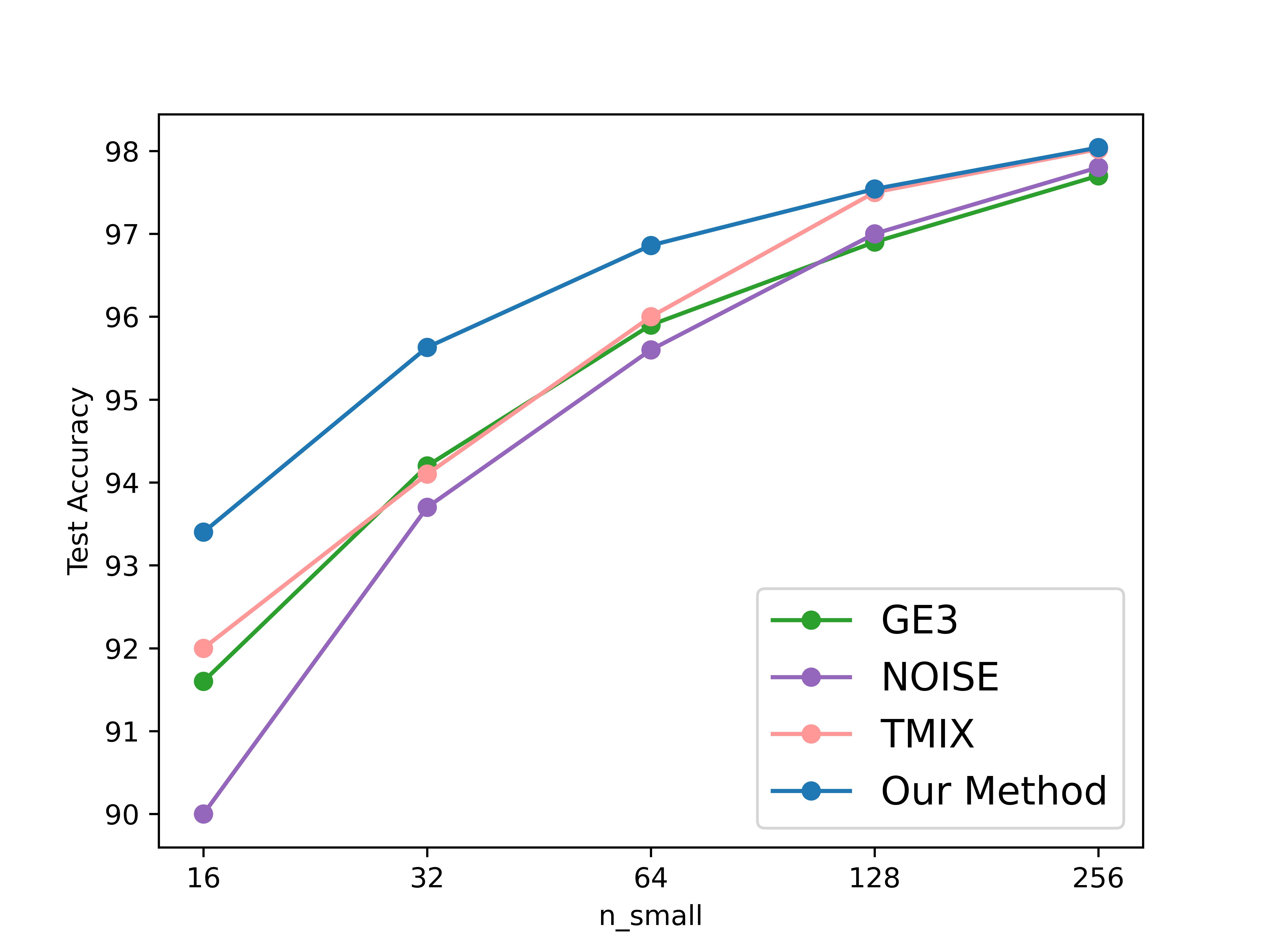}} 

\caption{The performance of \textsc{Reprint}, GE3, NOISE and TMIX on Yahoo Answer (left panel) and DBpedia (right panel) in broader class-imbalanced scenarios with $n_{\textrm{small}}$ in \{16, 32, 64, 128, 256\}.}
\label{fig:fig2}
\end{figure}

Further, we also consider two more choices for $n_{\textrm{small}}$ (\emph{i.e.,} $n_{\textrm{small}}\in\{16, 256\}$) to investigate the performance of our method in more severe imbalanced scenario and less severe one. Figure \ref{fig:fig2} shows the mean accuracy of both our method and other three competitive baseline models on Yahoo Answer and DBpedia datasets under the consideration of five $n_{\textrm{small}}$ values. It can be seen clearly that the improvement of our method over baselines is much more obvious when the training data is more imbalanced even though the estimation of principal components may be of large bias when the number of samples is small. In addition, as shown in Table \ref{tab:table1}, an average of 3.94\% (or 1.91\%) improvement is attained over GE3 in each scenario on Yahoo Answer (or AG news), which indicates the generality and stability of our method in different class-imbalanced settings.

\subsection{Ablation studies}
We now present ablation studies to show the effectiveness of subspace representation and label refinement in \textsc{Reprint}.

\subsubsection{Different strategies for choosing principal components}
We compare two strategies mentioned in Section \hyperref[sec:choice of pc]{3.2} to determine the subspace representation for source and target class:
\begin{enumerate}
    \item We consider $h=q=k$ and $k in \{1, 5, 10, 15\}$.
    \item We set the ratio of variance explained in \{50\%, 70\%, 95\%\}.
\end{enumerate}

\begin{figure}[t]
\centering

\subfigure{\includegraphics[scale=.5]{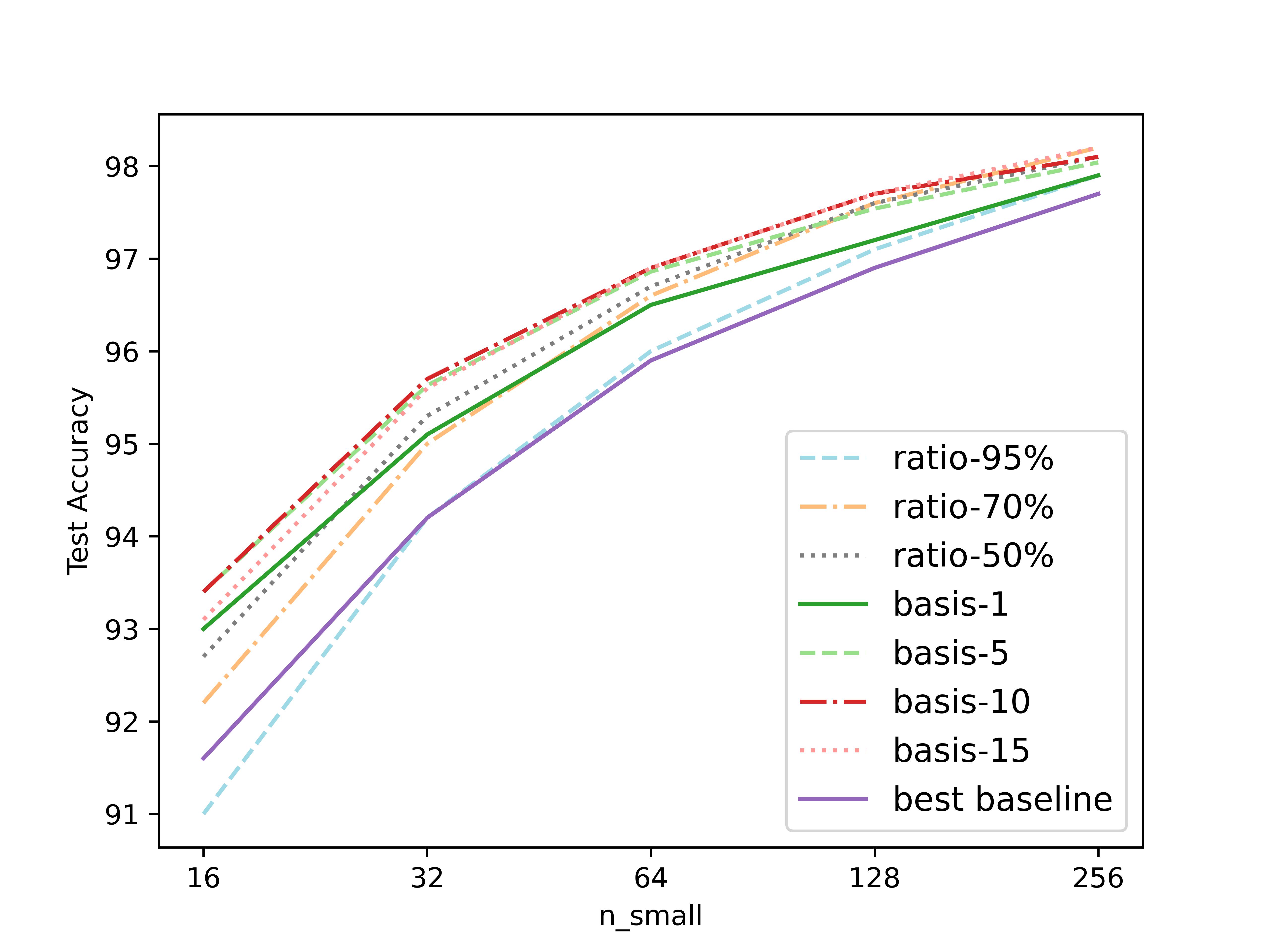}}
\subfigure{\includegraphics[scale=.5]{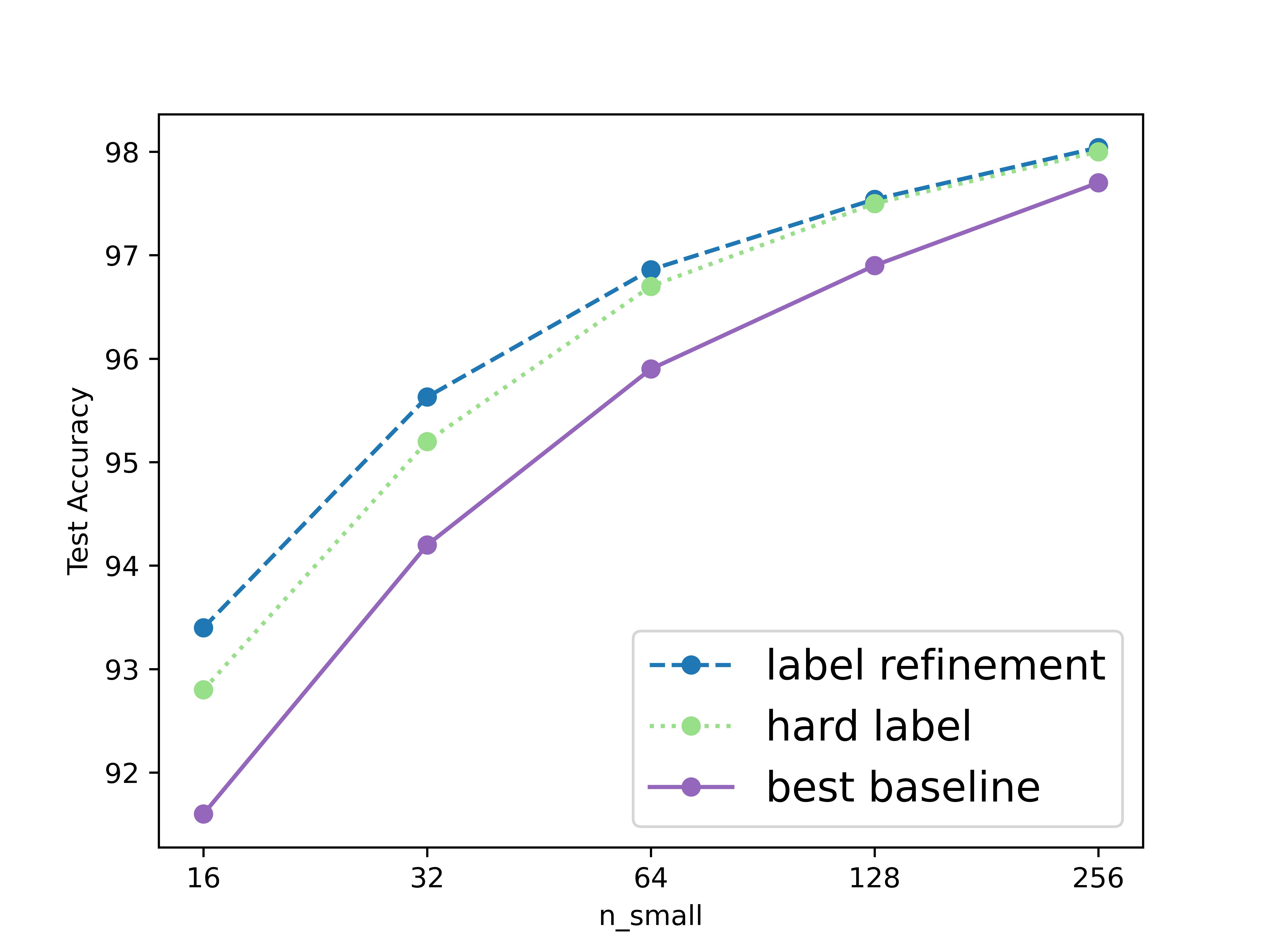}}
\caption{Ablation study of \textsc{Reprint}~on DBpedia. The left plot gives the performances with several choices of principal components while the right plot shows the performances of our method with two different choices of labels.}
\label{fig:fig3}
\end{figure}

The left panel of Figure \ref{fig:fig3} shows the test accuracy of \textsc{Reprint}~on DBpedia, using different strategies to choose principal components. We observe that the performance degrades when the ratio of explained variance increases from 50\% to 95\%, suggesting that only noises, rather than informative residual that relates to semantic and syntactic structure of source class, are incorporated in the generate augmented examples when the ratio comes to 95\%. Additionally, as 1, 5, 10 and 15 principal components approximate respectively to ratio of 20\%, 40\%, 45\% and 50\%, the comparatively poor performance of our method with 
one principal component (dark green line) also indicates that the performance would deteriorate if the number of principal components is too small. In other words, the subspace formed by principal components is too simple to provide sufficient representation for the underlying pattern of class distribution. On the contrary, as long as the ratio takes intermediate values, one can see that their performances always rank among the best even with various $n_{\textrm{small}}$ values and that their differences are small, suggesting that using subspace representation of class distribution when generating augmented examples helps to bring useful distributional information from one class to another, and this benefits the versatility of the target class. For this reason, we heuristically set $h=q=5$ for the following experiments.

\subsubsection{Different choices for labels}
The mixup-based approaches \citep{mixtext, ZCD2018} have shed light on our label refinement discussed in Section \hyperref[sec:label refinement]{3.3}. In this part, we would like to study how this refinement boosts the final result by carrying out ablation studies with the following two choices of labels for augmented examples.

\begin{enumerate}
    
    \item \textbf{Label refinement} creates new soft labels for augmented examples based on the principal components, as discussed in equation \eqref{eq:created label} 
    
    \item \textbf{Hard label} preserves labels for those augmented examples and represents them by one-hot vector.
\end{enumerate}

The right panel of Figure \ref{fig:fig3} reports the accuracy achieved using aforementioned choices of labels. It is obvious that the proposed label refinement can achieve better performance than hard label with various $n_\textrm{small}$ values, and this gap becomes more significant in more imbalanced scenario. We conjecture that label refinement component can serve as regularization effect to the  classification model and hence improve the model generalization ability.

\begin{table}[h]
\centering
\begin{tabular}{cc}
\hline
Strategy           & Accuracy(\%) \\ \hline
\textsc{Reprint}              & 93.4     \\
without label refinement       & 92.8     \\
without label refinement $\&$ subspace representation   & 91.6     \\ \hline
\end{tabular}
\vspace{0.7em}
\caption{Mean test accuracy of \textsc{Reprint}~when removing label refinement and subspace representation one by one. The results are obtained on DBpedia with $n_{\textrm{small}}=16$.}
\label{tab:table3}
\end{table}

To illustrate the contribution provided by the two components involved in our method, Table \ref{tab:table3} reports the decline of the accuracy of \textsc{Reprint}~on DBpedia dataset after striping label refinement and principal components-based subspace representation consecutively. It is noted that the drop is significantly larger when we remove subspace representation than only removing label refinement, indicating that the subspace representation that is used when extrapolating augmented examples has larger impact than the label refinement on the final performance of \textsc{Reprint}, which highlights the crucial role of using principal components in our method.

\subsection{Different hidden-space representation}
In addition, to demonstrate that \textsc{Reprint}~is not tailored for only one kind of hidden-space representations yielded by some certain language model but also has its generality and effectiveness with other model, we present the results in Table \ref{tab:table2} with hidden-space representations now given by another pretrained language model ALBERT (\citealp{LCG2020}). It is noted that our method achieves an average improvement of 4.69\% of accuracy over UPSPL, and also outweighs other baseline models.  

\begin{table}[ht]
\centering
\begin{tabular}{c|ccc|ccc}
\hline
Dataset      & \multicolumn{3}{c|}{Snips}   & \multicolumn{3}{c}{Yahoo Answer}    \\ \hline
$n_{\textrm{small}}$  & 32       & 64       & 128    & 32      & 64      & 128     \\ \hline
UPSPL   & $87.74_{\pm0.88}$     & $91.03_{\pm0.88}$     & $93.23_{\pm0.44}$   & $46.56_{\pm1.01}$   & $50.23_{\pm0.61}$     & $53.68_{\pm0.92}$    \\
SMOTE        & $87.77_{\pm0.69}$     & $91.03_{\pm0.71}$     & $93.03_{\pm0.51}$   & $46.01_{\pm0.60}$    & $49.93_{\pm0.77}$     & $53.51_{\pm0.94}$    \\
TMIX         & $88.80_{\pm1.80}$     & $92.06_{\pm1.28}$     & $93.31_{\pm0.93}$   & $57.45_{\pm2.15}$    & $58.05_{\pm1.23}$     & $60.88_{\pm1.57}$    \\
EDA          & $87.36_{\pm1.35}$     & $91.03_{\pm1.54}$     & $92.14_{\pm0.65}$   & $46.23_{\pm0.99}$    & $49.76_{\pm0.89}$     & $53.60_{\pm0.86}$    \\
LD           & $82.66_{\pm1.21}$     & $87.74_{\pm1.36}$     & $90.31_{\pm1.14}$   & $46.48_{\pm0.37}$    & $51.88_{\pm0.61}$     & $57.20_{\pm0.78}$    \\
WE           & $83.89_{\pm1.28}$     & $88.31_{\pm1.88}$     & $90.91_{\pm1.24}$   & $44.46_{\pm0.37}$    & $50.14_{\pm0.71}$     & $55.28_{\pm0.67}$    \\
NOISE        & $88.17_{\pm0.72}$     & $91.37_{\pm1.30}$     & $93.26_{\pm0.78}$   & $50.77_{\pm0.41}$    & $55.88_{\pm0.52}$     & $59.71_{\pm0.59}$     \\
GE3          & $90.00_{\pm0.69}$     & $92.51_{\pm0.99}$     & $94.09_{\pm0.59}$   & $52.68_{\pm0.41}$    & $57.50_{\pm0.64}$     & $61.06_{\pm0.63}$     \\
\textsc{Reprint}         & $\bm{90.83_{\pm0.69}}$     & $\bm{92.74_{\pm0.94}}$     & $\bm{94.29_{\pm0.80}}$   & $\bm{57.98_{\pm0.5}}$    & $\bm{61.29_{\pm0.36}}$    &    $\bm{62.99_{\pm0.52}}$     \\ \hline
             & \multicolumn{3}{c|}{AG news} & \multicolumn{3}{c}{DBpedia} \\ \hline
UPSPL   & $72.44_{\pm2.27}$     & $76.79_{\pm0.89}$     & $80.39_{\pm0.60}$   & $90.01_{\pm0.36}$    & $92.97_{\pm0.37}$    & $94.97_{\pm0.27}$    \\
SMOTE        & $72.55_{\pm2.20}$     & $76.91_{\pm0.83}$     & $80.49_{\pm0.66}$   & $89.85_{\pm0.39}$    & $92.78_{\pm0.40}$    & $94.82_{\pm0.29}$    \\
TMIX         & $74.26_{\pm1.64}$     & $78.97_{\pm1.27}$     & $82.36_{\pm1.01}$   & $90.86_{\pm1.37}$    & $94.59_{\pm1.25}$    & $95.20_{\pm1.13}$    \\
EDA          & $71.26_{\pm1.29}$     & $77.20_{\pm1.25}$     & $80.39_{\pm0.67}$   & $90.32_{\pm1.76}$    & $91.70_{\pm1.13}$    & $94.13_{\pm1.04}$    \\
LD           & $65.69_{\pm1.08}$     & $69.28_{\pm0.69}$     & $73.37_{\pm0.79}$   & $82.97_{\pm0.64}$    & $87.21_{\pm0.35}$    & $91.20_{\pm0.36}$    \\
WE           & $64.80_{\pm1.69}$     & $68.98_{\pm0.83}$     & $72.62_{\pm0.75}$   & $83.28_{\pm0.65}$    & $87.91_{\pm0.49}$    & $91.47_{\pm0.17}$    \\
NOISE        & $73.50_{\pm2.16}$     & $77.98_{\pm0.80}$     & $81.68_{\pm0.53}$   & $90.37_{\pm0.27}$    & $93.25_{\pm0.34}$    & $95.13_{\pm0.26}$    \\
GE3          & $76.08_{\pm0.94}$     & $79.68_{\pm0.83}$     & $82.93_{\pm0.34}$   & $92.24_{\pm0.45}$    & $94.32_{\pm0.20}$    & $95.61_{\pm0.09}$    \\
\textsc{Reprint}         & $\bm{76.71_{\pm1.30}}$     & $\bm{81.13_{\pm1.04}}$     & $\bm{84.50_{\pm0.54}}$   & $\bm{92.97_{\pm0.30}}$    & $\bm{94.92_{\pm0.24}}$    & $\bm{95.94_{\pm0.14}}$    \\ \hline
\end{tabular}
\vspace{0.7em}
\caption{Results using ALBERT pretrained language model}
\label{tab:table2}
\end{table}

\section{Conclusion and future work}\label{sec:future work}

To alleviate data scarcity or class imbalance, \textsc{Reprint}~provides a new perspective for hidden-space data augmentation. To generate augmented samples with both fidelity and variety to the target class distribution, we make use of principal components to form subspace representations for both source and target classes in a lower dimension, thus its primary semantic and syntactic structure are preserved. Served as a variant containing additional information, the difference between original sample and its subspace representation are then used to increase the diversity of augmented samples for the target. Moreover, we provide an alongside label refinement component enabling the creation of new soft labels for augmented examples which can boost the performance of our method. Through experiments on four text classification benchmark datasets, we illustrate the effectiveness of \textsc{Reprint}~which has better test accuracy and competitive standard deviation when compared with other eight data augmentation methods in NLP. Moreover, \textsc{Reprint}~is appealing since it demands low computational resource and the only hyperparameter contained is the one determining the dimension of subspace.

Additionally, we provide ablation studies showing the effect of both subspace representation and label refinement components on the final results. It is found that when the explained variance ratio of principal components is of medium values, the subspace representation formed by them improves the most to the test accuracy and the improvement is stable in several class-imbalanced scenarios. Furthermore, it is also found that label refinement that is involved in our method achieves better results in comparison with hard label which preserves labels for augmented examples.

In the future, we would like to extend the characterization of the geometry of class distribution from linear representation to non-linear setting, either by using principal curves or other manifold learning methods, for the purpose of providing a more elaborate description of class distribution. In addition, we plan to explore whether our method can be adapted to other settings such as semi-supervised learning. Finally, we also seek to ameliorate the sampling distribution by considering the importance of each target example in terms of their gradients or other metrics.  

\section*{Acknowledgments}
This work is supported by National Natural Science Foundation of China (Grant No. 62107021) and Hubei Provincial Science and Technology Innovation Base (Platform) Special Project 2020DFH002.

\bibliographystyle{plainnat}  
\bibliography{references}

\end{document}